\documentclass{article}
\usepackage{spconf,amsmath,graphicx}
\usepackage{times}
\usepackage{epsfig}
\usepackage{graphicx}
\usepackage{amsmath}
\usepackage{amssymb}
\usepackage{adjustbox}
\usepackage{tabularx}

\usepackage{subfigure}

\usepackage{multirow}
\usepackage{textcomp}
\usepackage{colortbl}
\usepackage{xcolor}
\usepackage{array,tabularx}
\usepackage{enumitem}
\usepackage{adjustbox}

\title{Semantic Segmentation and Object Detection Towards Instance Segmentation: Breast Tumor Identification}
\name{Mohamed Mejri$^{1}$,Aymen Mejri $^{2}$ Oumayma Mejri$^{3}$ Chiraz Fekih$^{4}$}
\address{School of Electrical and Computer Engineering, Georgia Institute of Technology
, USA$^{1}$\\Department of digital signal processing, Telecom Paris, Paris, France$^{2}$ \\Department of gynecology and obstetrics, Medicine School of Tunis, Tunis, Tunisia$^{3,4}$}

\begin{document}
%

\maketitle
\begin{abstract}
Breast cancer is one of the factors that cause the increase of mortality of women. The most widely used method for diagnosing this geological disease i.e. breast cancer is the ultrasound scan. Several key features such as the smoothness and the texture of the tumor captured through ultrasound scans encode the abnormality of the breast tumors (malignant from benign). However, ultrasound scans are often noisy and include irrelevant parts of the breast that may bias the segmentation of eventual tumors. In this paper, we are going to extract the region of interest ( i.e, bounding boxes of the tumors) and feed-forward them to one semantic segmentation encoder-decoder structure based on its classification (i.e, malignant or benign). the whole process aims to build an instance-based segmenter from a semantic segmenter and an object detector.
\end{abstract}
\section{Introduction}
Over the past few years, cancers were classified as the disease of the century killing hundreds of thousands of persons. Breast cancer is among the most common causes of cancer deaths today, coming fifth after lung, stomach, liver, and colon cancers. Although, early diagnosis through ultrasound scan helps mitigate its mortality, analyzing the key features of the ultrasound scan is still challenging. 
Hence, the medical and vision community collaborated to automate breast cancer diagnosis through machine learning techniques. In this context, we are introducing breast tumors ultrasound segmentation (i.e, malignant and benign classes) through an instance-based segmentation approach to detect and segment tumors at once.

\section{Related work}
In this section, we will detail the state of the art of instance-based segmentation as well as the work done in breast cancer segmentation.
\subsection{Instance segmentation neural networks}
Given its importance, a lot of research effort has been made to push instance segmentation accuracy. Mask-RCNN\cite{he2018mask} is a two-stage instance-based segmentation model that first generates candidates' region of interests (ROIs) and then classifies and segments those ROIs in the second stage. Other related works try to improve its accuracy by enriching the FPN\cite{lin2017feature} features. This model requires re-pooling features for each ROI and hence are computationally more expensive. Finally, other algorithms first perform semantic segmentation followed by boundary detection\cite{kirillov2016instancecut}, pixel clustering\cite{bai2017deep,liang2015proposalfree} or learn an embedding to form instance masks\cite{newell2017associative,harley2017segmentationaware,debrabandere2017semantic,fathi2017semantic}.

While real time object detection \cite{bochkovskiy2020yolov4,redmon2016look,redmon2016yolo9000,Liu_2016,redmon2018yolov3} and semantic segmentation methods exist, few work have been done to deal with real time instance segmentation. Straight to shape\cite{jetley2017straight} and Box2Pix \cite{UB18} can perform instance segmentation in real-time (30 fps on Pascal SBD 2012 \cite{everingham2010pascal,6126343} for Straight to Shapes, and 10.9 fps on Cityscapes \cite{cordts2016cityscapes} and 35 fps on KITTI \cite{6248074} for Box2Pix) but their performance are far from that of modern baselines.

\subsection{Breast cancer segementation}
\cite{PMID:31867417} introduced a new dataset collected at baseline that includes breast ultrasound images among women between ages between 25 and 75 years old. This data was collected in 2018. The number of patients is 600 female patients. The dataset consists of 780 images with an average image size of 500×500 pixels. The images were obtained using LOGIQ E9 ultrasound system and LOGIQ E9 Agile ultrasound system. Their transducers are 1e5 MHz on ML6-15-D Matrix linear probe. They are categorized into three classes: normal, benign, and malignant. While \cite{MOON2020105361,BYRA2020102027,mohammadi2021ultrasound,gomez2020comparative,jimenez2020deep} worked on novel methods to segment and/or classify those images, \cite{MOON2020105361} uses an ensemble-based strategy to classify the breast ultrasound scans: Four classifiers were trained (i.e, a combination of VGGNet\cite{simonyan2015deep}, ResNet\cite{he2015deep}, DenseNet\cite{huang2018densely}) on four different image versions to predict the category (benign, malignant) of potential masses on the ultrasound image.\cite{BYRA2020102027} proposed a selective kernel U-Net based on an attention mechanism. The SK-U-Net achieved a mean Dice score of 0.826 and outperformed the regular U-Net\cite{ronneberger2015unet}, Dice score of 0.778.     

\section{Methods}
In this section, we will present a whole pipeline for tumors detection and segmentation.
\subsection{Tumor detection pipeline}
\begin{figure}[ht]
\centerline{\includegraphics[width=9cm,height=12cm,keepaspectratio]{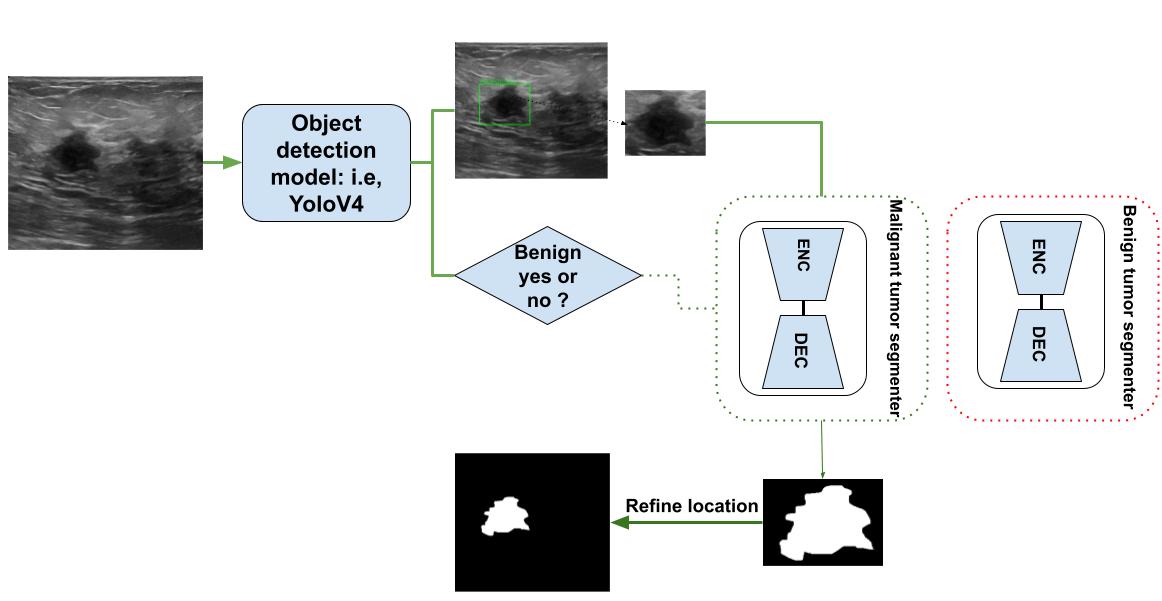}}
\caption{Pipeline for breast tumor segmentation}
\label{fig1}
\end{figure}

The pipeline consists of a super-resolution resizing preprocess followed by object detection algorithms that identify the tumors' location as well as their abnormality form (malignant or benign). We choose EfficientDet\cite{tan2020efficientdet} as an object detector since it shows high detection performance on common datasets (MS-COCO\cite{lin2015microsoft} and PASCAL-VOC\cite{everingham2010pascal}). Then each tumor is cropped following its bounding box, resized using a super-resolution process, and feed-forward to an encoder-decoder segmenter depending on its class. Finally, the predicted tumor location is refined through the detected anomaly bounding box coordinates.
The figure \ref{fig1} detailed the pipeline steps.
Although anomaly detection inside breast ultrasound scan is a key step for breast cancer prediction, semantic segmentation of the tumor itself remains important since it shows the key features and characteristics of the tumor shape. However, due to the shortage of the dataset\cite{mejri2020randomforestmlp}, a data augmentation pre-processing is necessary.

\section{Experiments and Results}
In this section, we are going to discuss different experiments done with the tumor detection pipeline components.
\subsection{Breast cancer segmentation dataset}
It consists of 778 ultrasound scan images with their corresponding masks: 435 images containing benign breast tumors, 210 images of malignant breast tumors, and 133 normal breast images. The bounding boxes are automatically extracted from the mask. We added an offset of 10 pixels to each bounding box width and height to capture appropriately the contrast between the edges of the tumor and the normal tissue of the breast. We augmented the dataset 6 times through basic and deep learning-based data augmentation techniques. 
The figures below show different ultrasound scan images with their corresponding mask (red area).
The dataset was split into a train set (80\% of the dataset), a validation set (10\% of the dataset), and a test set (10\% of the dataset). we choose to perform cross-validation to mitigate the effect of outliers: We split the train-validation set into 9 folds based on their classes (benign, malignant, or normal). Each time 8 folds are used as a train set and the remaining folder is used for validation.    
\begin{figure}[htbp]
\centerline{\includegraphics[width=8cm,height=3cm]{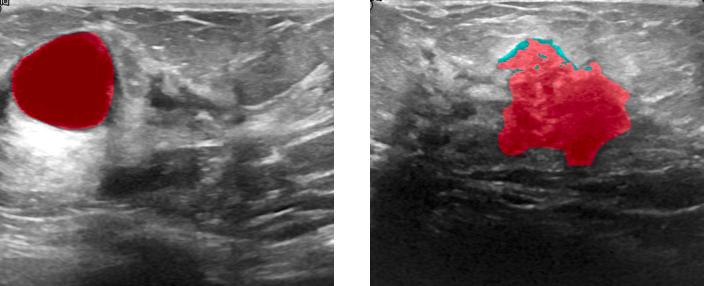}}
\caption{(Right) breast malignant tumor (Left) breast benign tumor}
\label{fig}
\end{figure}
\subsection{Object Detection Algorithm}
The graph below shows the mean average precision of different object detection algorithms as well as their inference time when applied to the breast cancer ultrasound scan dataset. Each model is trained for 50 epochs. At the end of each
epoch, a validation test is performed, and we save only the weights of the model with a maximum validation mAP. All experiments were carried out on an IBM Power Systems AC922 with 256 GB of RAM and 4 NVIDIA V100 16 GB GPGPU (using only a single GPGPU).
\begin{figure}[htbp]
\centerline{\includegraphics[width=8cm,height=8cm]{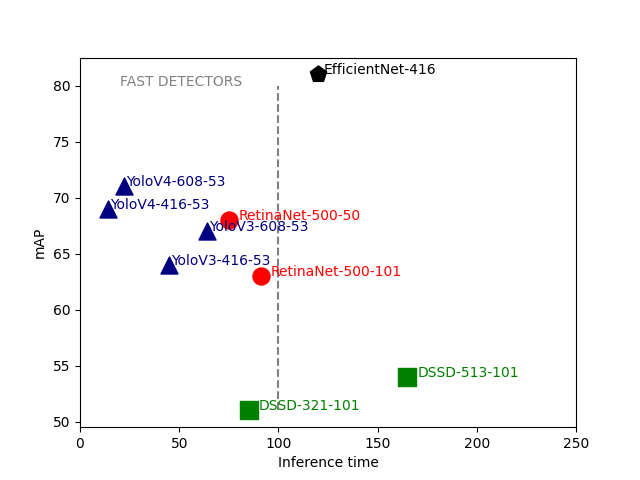}}
\caption{Object detection mAP and inference time}
\label{obj}
\end{figure}
In the graph\ref{obj}, the object detectors are denoted as following (object detector, image size, backbone size) and the whole mAP-inference time-space is split into fast and slow detectors following the limit of 100 in inference time. Many models and especially the YOLO\cite{bochkovskiy2020yolov4}-based ones are at the real-time pace detection (25 FPS). The analysis of the breast ultrasound scans is done prospectively and the real-time constraint is not applicable and only the mean average precision is the key metric for detection. However, even the slowest detector could be accelerated on GPU using tensorRT\cite{8844665} and we gain 40 ms in inference time for the EfficientDet-416\cite{tan2020efficientdet} model for example. According to the graph we choose the EfficientDet-416\cite{tan2020efficientdet} as the best object detector from the mAP point of view. A non-maxima-suppression post-process is necessary to remove irrelevant bounding boxes. We chose the confidence score threshold to be equal to 0.6 and the intersection over the union between neighbor bounding boxes as 0.4.   
In the following section, we will detail the experiments done on semantic segmentation models
\subsection{Semantic Segmentation Algorithms}
Several semantic segmentation algorithms have been assessed to predict the tumors inside the region of interest captured through object detection algorithms. 
Two DeepLabV3plus\cite{chen2018encoderdecoder} was trained for 50 epochs on 80\% of the dataset on a cross-validation basis to avoid overfitting: one the malignant based breast ultrasound images subset and another on the benign tumors. Hence, we will assess two predictors instead of one. To capture appropriately the edges of the tumor especially the malignant class, we chose to use The Dice-loss\cite{wang2018focal}. 

\begin{equation}
\label{CE}
\setlength\abovedisplayskip{19pt}
\setlength\belowdisplayskip{19pt}
 \mathcal{L} = -\tfrac{\sum_{i}^{N} p_{i}g_{i}}{\sum_{i}^{N}p_{i}^{2}+\sum_{i}^{N}g_{i}^{2}}.
\end{equation}

where $p_{i}$ refers to the predicted probabilities and $g_{i}$ to the ground truth. \newline
The hyperparameter of the predictor is tuned as follows: we chose an Adam optimizer with an initial learning rate of $10^{-3}$, a learning rate scheduler is also used to avoid undesirable divergent behavior while increasing the number of epochs(the learning rate drops to $10^{-4}$ after 20 epochs).

The table below shows the root mean squared error (RMSE) of different algorithms.
\begin{table}[htp]
\caption{\centering RMSE of the different type of architectures : On breast cancer segmentation dataset. Lower is better.} \label{tab:SRMSE}
\begin{center}
\scalebox{0.6}{
\begin{tabular}{|c|c|c|c|c|c|c|}
\hline
               & DeepLabV3plus\cite{chen2018encoderdecoder} & DeepLabV3\cite{chen2017rethinking} & U-Net\cite{ronneberger2015unet} & SegNet\cite{badrinarayanan2016segnet} & FCN\cite{long2015fully}  & PSPNet\cite{zhao2017pyramid} \\ \hline
RMSE malignant & \textbf{0.11}         & 0.15      & 0.23  & 0.18   & 0.19 & 0.17   \\ \hline
RMSE benign    & \textbf{0.08}          & 0.12      & 0.18  & 0.14   & 0.13 & 0.15   \\ \hline
\end{tabular}
}
\end{center}
\label{tab1}
\end{table}
According to the table above the DeepLabV3plus\cite{chen2018encoderdecoder} is the best predictor for both malignant and benign tumors. However, the RMSE of the predicted malignant tumors is higher than the RMSE of the benign tumors. This observation is expected since the malignant tumor has a very complex shape compared to benign tumors and hence is more difficult to segment.   Many backbones exist and could be used as feature extractors in DeepLabV3plus\cite{chen2018encoderdecoder}. The table below shows the RMSE of different DeepLabV3plus\cite{chen2018encoderdecoder} version using different backbones.

\begin{table}[htp]
\begin{center}
\caption{\centering RMSE and parameter number of different backbones (encoder of DeepLabV3plus\cite{chen2018encoderdecoder})} \label{tab:SRMSE}
\scalebox{0.65}
{\begin{tabular}{|c|c|c|c|c|c|}
\hline
                 & ResNet-101\cite{he2015deep} & InceptionV3\cite{szegedy2015rethinking} & Xception\cite{chollet2017xception} & VGG16\cite{simonyan2015deep} & VGG19\cite{simonyan2015deep} \\ \hline
RMSE malignant   & 0.16       & 0.19        & \textbf{0.11}     & 0.2   & 0.21  \\ \hline
RMSE benign      & 0.11       & 0.12        & \textbf{0.08}     & 0.16  & 0.17  \\ \hline
Parameter number & 23M        & 24M         & \textbf{22M}      & 138M  & 144M  \\ \hline
\end{tabular}
}
\end{center}
\end{table}

The Xception structure used as a backbone achieves better performance when it comes to segmenting breast both malignant and benign tumors with DeepLabV3plus\cite{chen2018encoderdecoder} as expected. Xception is also more robust to overfitting compared to other neural networks thanks to its low number of parameters.
As we have assessed all the instance segmentation pipeline components, let's analyze the outcome of the whole instance segmentation method.
\subsection{Instance segmentation pipeline}
As stated in the previous section, our instance segmentation pipeline consists of an object detection algorithm to extract region of interest and multiple semantic segmentation models. Each one represent a class. The figures below shows a comparison of outcomes between the commonly used Mask-RCNN\cite{he2018mask} and our model which are pre-trained on PASCAL-VOC\cite{everingham2010pascal}. 
\begin{figure}[htbp]
\centerline{\includegraphics[width=8cm,height=5cm]{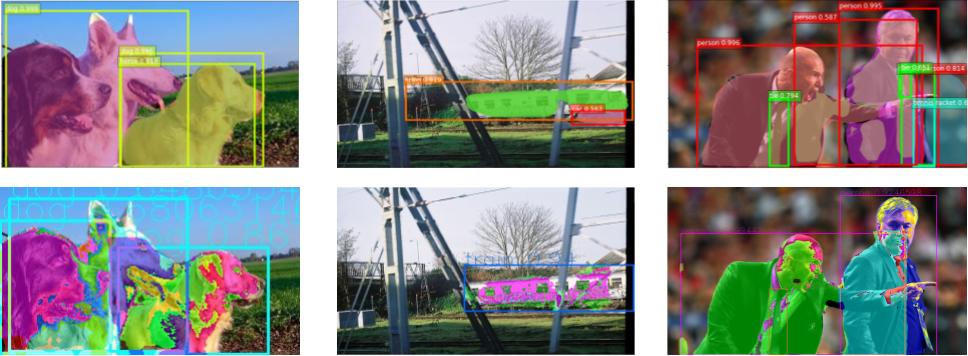}}
\caption{Object detection mAP and inference time}
\label{benign}
\end{figure}
The figures below show the advantages and the limitation of our model in instance segmentation tasks. The Mask-RCNN\cite{he2018mask} couldn't capture the three dogs (top-left) and only segment two dogs as one and misclassify the remaining one as a horse while the three dogs are predicted appropriately but not well segmented due to occlusion inside the image. similarly for the top right prediction where the two men are well-segmented using our model. All in all, the Mask-RCNN\cite{he2018mask} makes a lot of false alarms while our model predicts well the bounding boxes of objects. One limitation of our model is its lack of robustness to occlusion and hence it is not appropriate for predicting the mask of overlapping objects. 
Despite that, it is very efficient when it comes to segmenting the breast tumors since they are geologically not overlapping. The figures below shows two breast ultrasound scans with their corresponding predicted tumors (i.e, blue mask).
\begin{figure}[ht]
\centerline{\includegraphics[width=6cm,height=4cm]{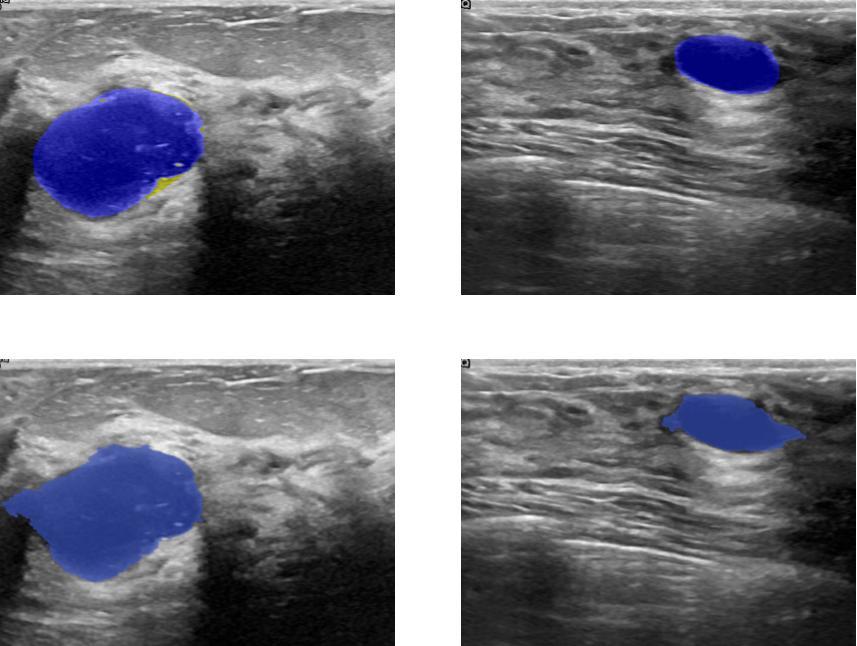}}
\caption{Two benign tumors ultrasound scan: (Upper-left): $1^{st}$Ground truth (Upper-right): $2^{nd}$ Ground truth $1^{st}$ Ground truth: (Lower-left) $1^{st}$ Prediction (Lower-right): $2^{nd}$ Prediction }
\label{malignant}
\end{figure}
Benign tumors\cite{levy2007birads} share a number of features on ultrasound that indicate their benignity. In particular a round or oval shape , a large axis parallel to the skin, circumscribed margins or gently curving smooth lobulations , hyperechoic tissue in the case of solid tumors and anechoic content in the case of cysts and acoustic shadows in the edge. In contrast, malignant tumors\cite{levy2007birads} show specific features on ultrasound including, irregular shape with microlobulations or spiculations, a large axis that is not parallel to the skin, poorly defined margins, hypoechoic nodule or hetergenous echostructure and acoustic posterior shadowing. a visual analysis of the figures \ref{benign} and \ref{malignant} prove that our model is able to capture the most relevant features we have already described.  

\begin{figure}[ht]
\centerline{\includegraphics[width=6cm,height=4cm]{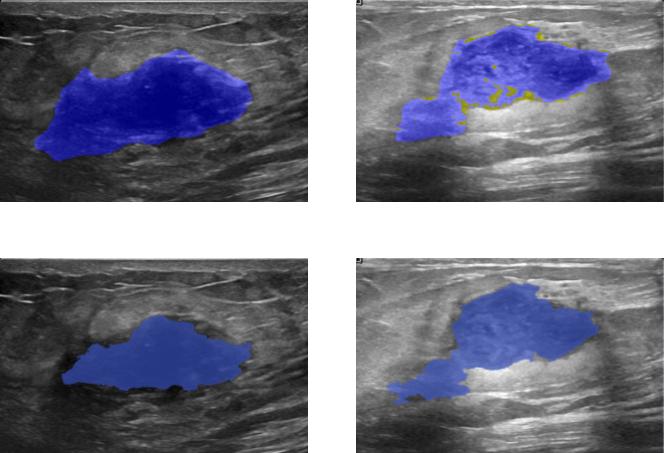}}
\caption{Two malignant tumors ultrasound scan: (Upper-left): $1^{st}$Ground truth (Upper-right): $2^{nd}$ Ground truth $1^{st}$ Ground truth: (Lower-left) $1^{st}$ Prediction (Lower-right): $2^{nd}$ Prediction }\label{fig}
\end{figure}
\section{Conclusion}
Artificial intelligence has been increasingly applied in ultrasound and proved to be a powerful tool to provide a reliable diagnosis with higher accuracy and efficiency and reduce the workload of physicians. In this paper, we introduced a novel application of instance segmentation techniques on breast tumors segmentation. We provided evidence of the inefficiency of the classical instance segmentation model and we build our algorithm to detect and segment malignant and benign tumors at the same time. Further work must be done to adapt the proposed algorithm to fit overlapping objects and to be more robust to occlusion.

\label{sec:ref}

\bibliographystyle{IEEEbib}
\bibliography{refs}

\begin{thebibliography}{10}

\bibitem{he2018mask}
Kaiming He, Georgia Gkioxari, Piotr Dollár, and Ross Girshick,
\newblock ``Mask r-cnn,'' 2018.

\bibitem{lin2017feature}
Tsung-Yi Lin, Piotr Dollár, Ross Girshick, Kaiming He, Bharath Hariharan, and
  Serge Belongie,
\newblock ``Feature pyramid networks for object detection,'' 2017.

\bibitem{kirillov2016instancecut}
Alexander Kirillov, Evgeny Levinkov, Bjoern Andres, Bogdan Savchynskyy, and
  Carsten Rother,
\newblock ``Instancecut: from edges to instances with multicut,'' 2016.

\bibitem{bai2017deep}
Min Bai and Raquel Urtasun,
\newblock ``Deep watershed transform for instance segmentation,'' 2017.

\bibitem{liang2015proposalfree}
Xiaodan Liang, Yunchao Wei, Xiaohui Shen, Jianchao Yang, Liang Lin, and
  Shuicheng Yan,
\newblock ``Proposal-free network for instance-level object segmentation,''
  2015.

\bibitem{newell2017associative}
Alejandro Newell, Zhiao Huang, and Jia Deng,
\newblock ``Associative embedding: End-to-end learning for joint detection and
  grouping,'' 2017.

\bibitem{harley2017segmentationaware}
Adam~W. Harley, Konstantinos~G. Derpanis, and Iasonas Kokkinos,
\newblock ``Segmentation-aware convolutional networks using local attention
  masks,'' 2017.

\bibitem{debrabandere2017semantic}
Bert~De Brabandere, Davy Neven, and Luc~Van Gool,
\newblock ``Semantic instance segmentation with a discriminative loss
  function,'' 2017.

\bibitem{fathi2017semantic}
Alireza Fathi, Zbigniew Wojna, Vivek Rathod, Peng Wang, Hyun~Oh Song, Sergio
  Guadarrama, and Kevin~P. Murphy,
\newblock ``Semantic instance segmentation via deep metric learning,'' 2017.

\bibitem{bochkovskiy2020yolov4}
Alexey Bochkovskiy, Chien-Yao Wang, and Hong-Yuan~Mark Liao,
\newblock ``Yolov4: Optimal speed and accuracy of object detection,'' 2020.

\bibitem{redmon2016look}
Joseph Redmon, Santosh Divvala, Ross Girshick, and Ali Farhadi,
\newblock ``You only look once: Unified, real-time object detection,'' 2016.

\bibitem{redmon2016yolo9000}
Joseph Redmon and Ali Farhadi,
\newblock ``Yolo9000: Better, faster, stronger,'' 2016.

\bibitem{Liu_2016}
Wei Liu, Dragomir Anguelov, Dumitru Erhan, Christian Szegedy, Scott Reed,
  Cheng-Yang Fu, and Alexander~C. Berg,
\newblock ``Ssd: Single shot multibox detector,''
\newblock {\em Lecture Notes in Computer Science}, p. 21–37, 2016.

\bibitem{redmon2018yolov3}
Joseph Redmon and Ali Farhadi,
\newblock ``Yolov3: An incremental improvement,'' 2018.

\bibitem{jetley2017straight}
Saumya Jetley, Michael Sapienza, Stuart Golodetz, and Philip H.~S. Torr,
\newblock ``Straight to shapes: Real-time detection of encoded shapes,'' 2017.

\bibitem{UB18}
J.~Uhrig, E.~Rehder, B.~Fr{\"o}hlich, U.~Franke, and T.~Brox,
\newblock ``Box2pix: Single-shot instance segmentation by assigning pixels to
  object boxes,''
\newblock in {\em IEEE Intelligent Vehicles Symposium (IV)}, 2018.

\bibitem{everingham2010pascal}
Mark Everingham, Luc Van~Gool, Christopher~KI Williams, John Winn, and Andrew
  Zisserman,
\newblock ``The pascal visual object classes (voc) challenge,''
\newblock {\em International journal of computer vision}, vol. 88, no. 2, pp.
  303--338, 2010.

\bibitem{6126343}
Bharath Hariharan, Pablo Arbeláez, Lubomir Bourdev, Subhransu Maji, and
  Jitendra Malik,
\newblock ``Semantic contours from inverse detectors,''
\newblock in {\em 2011 International Conference on Computer Vision}, 2011, pp.
  991--998.

\bibitem{cordts2016cityscapes}
Marius Cordts, Mohamed Omran, Sebastian Ramos, Timo Rehfeld, Markus Enzweiler,
  Rodrigo Benenson, Uwe Franke, Stefan Roth, and Bernt Schiele,
\newblock ``The cityscapes dataset for semantic urban scene understanding,''
  2016.

\bibitem{6248074}
Andreas Geiger, Philip Lenz, and Raquel Urtasun,
\newblock ``Are we ready for autonomous driving? the kitti vision benchmark
  suite,''
\newblock in {\em 2012 IEEE Conference on Computer Vision and Pattern
  Recognition}, 2012, pp. 3354--3361.

\bibitem{PMID:31867417}
Walid Al-Dhabyani, Mohammed Gomaa, Hussien Khaled, and Aly Fahmy,
\newblock ``Dataset of breast ultrasound images,''
\newblock {\em Data in brief}, vol. 28, pp. 104863, February 2020.

\bibitem{MOON2020105361}
Woo~Kyung Moon, Yan-Wei Lee, Hao-Hsiang Ke, Su~Hyun Lee, Chiun-Sheng Huang, and
  Ruey-Feng Chang,
\newblock ``Computer‐aided diagnosis of breast ultrasound images using
  ensemble learning from convolutional neural networks,''
\newblock {\em Computer Methods and Programs in Biomedicine}, vol. 190, pp.
  105361, 2020.

\bibitem{BYRA2020102027}
Michal Byra, Piotr Jarosik, Aleksandra Szubert, Michael Galperin, Haydee
  Ojeda-Fournier, Linda Olson, Mary O’Boyle, Christopher Comstock, and
  Michael Andre,
\newblock ``Breast mass segmentation in ultrasound with selective kernel u-net
  convolutional neural network,''
\newblock {\em Biomedical Signal Processing and Control}, vol. 61, pp. 102027,
  2020.

\bibitem{mohammadi2021ultrasound}
Narges Mohammadi, Marvin~M Doyley, and Mujdat Cetin,
\newblock ``Ultrasound elasticity imaging using physics-based models and
  learning-based plug-and-play priors,''
\newblock in {\em ICASSP 2021-2021 IEEE International Conference on Acoustics,
  Speech and Signal Processing (ICASSP)}. IEEE, 2021, pp. 1165--1169.

\bibitem{gomez2020comparative}
Wilfrido G{\'o}mez-Flores and Wagner~Coelho de~Albuquerque~Pereira,
\newblock ``A comparative study of pre-trained convolutional neural networks
  for semantic segmentation of breast tumors in ultrasound,''
\newblock {\em Computers in Biology and Medicine}, vol. 126, pp. 104036, 2020.

\bibitem{jimenez2020deep}
Yuliana Jim{\'e}nez-Gaona, Mar{\'\i}a~Jos{\'e} Rodr{\'\i}guez-{\'A}lvarez, and
  Vasudevan Lakshminarayanan,
\newblock ``Deep-learning-based computer-aided systems for breast cancer
  imaging: A critical review,''
\newblock {\em Applied Sciences}, vol. 10, no. 22, pp. 8298, 2020.

\bibitem{simonyan2015deep}
Karen Simonyan and Andrew Zisserman,
\newblock ``Very deep convolutional networks for large-scale image
  recognition,'' 2015.

\bibitem{he2015deep}
Kaiming He, Xiangyu Zhang, Shaoqing Ren, and Jian Sun,
\newblock ``Deep residual learning for image recognition,'' 2015.

\bibitem{huang2018densely}
Gao Huang, Zhuang Liu, Laurens van~der Maaten, and Kilian~Q. Weinberger,
\newblock ``Densely connected convolutional networks,'' 2018.

\bibitem{ronneberger2015unet}
Olaf Ronneberger, Philipp Fischer, and Thomas Brox,
\newblock ``U-net: Convolutional networks for biomedical image segmentation,''
  2015.

\bibitem{tan2020efficientdet}
Mingxing Tan, Ruoming Pang, and Quoc~V. Le,
\newblock ``Efficientdet: Scalable and efficient object detection,'' 2020.

\bibitem{lin2015microsoft}
Tsung-Yi Lin, Michael Maire, Serge Belongie, Lubomir Bourdev, Ross Girshick,
  James Hays, Pietro Perona, Deva Ramanan, C.~Lawrence Zitnick, and Piotr
  Dollár,
\newblock ``Microsoft coco: Common objects in context,'' 2015.

\bibitem{mejri2020randomforestmlp}
Mohamed Mejri and Aymen Mejri,
\newblock ``Randomforestmlp: An ensemble-based multi-layer perceptron against
  curse of dimensionality,'' 2020.

\bibitem{8844665}
Erik Smistad, Andreas Østvik, and André Pedersen,
\newblock ``High performance neural network inference, streaming, and
  visualization of medical images using fast,''
\newblock {\em IEEE Access}, vol. 7, pp. 136310--136321, 2019.

\bibitem{chen2018encoderdecoder}
Liang-Chieh Chen, Yukun Zhu, George Papandreou, Florian Schroff, and Hartwig
  Adam,
\newblock ``Encoder-decoder with atrous separable convolution for semantic
  image segmentation,'' 2018.

\bibitem{wang2018focal}
Pei Wang and Albert~CS Chung,
\newblock ``Focal dice loss and image dilation for brain tumor segmentation,''
\newblock in {\em Deep Learning in Medical Image Analysis and Multimodal
  Learning for Clinical Decision Support}, pp. 119--127. Springer, 2018.

\bibitem{chen2017rethinking}
Liang-Chieh Chen, George Papandreou, Florian Schroff, and Hartwig Adam,
\newblock ``Rethinking atrous convolution for semantic image segmentation,''
  2017.

\bibitem{badrinarayanan2016segnet}
Vijay Badrinarayanan, Alex Kendall, and Roberto Cipolla,
\newblock ``Segnet: A deep convolutional encoder-decoder architecture for image
  segmentation,'' 2016.

\bibitem{long2015fully}
Jonathan Long, Evan Shelhamer, and Trevor Darrell,
\newblock ``Fully convolutional networks for semantic segmentation,'' 2015.

\bibitem{zhao2017pyramid}
Hengshuang Zhao, Jianping Shi, Xiaojuan Qi, Xiaogang Wang, and Jiaya Jia,
\newblock ``Pyramid scene parsing network,'' 2017.

\bibitem{szegedy2015rethinking}
Christian Szegedy, Vincent Vanhoucke, Sergey Ioffe, Jonathon Shlens, and
  Zbigniew Wojna,
\newblock ``Rethinking the inception architecture for computer vision,'' 2015.

\bibitem{chollet2017xception}
François Chollet,
\newblock ``Xception: Deep learning with depthwise separable convolutions,''
  2017.

\bibitem{levy2007birads}
L~Levy, M~Suissa, JF~Chiche, G~Teman, and B~Martin,
\newblock ``Birads ultrasonography,''
\newblock {\em European journal of radiology}, vol. 61, no. 2, pp. 202--211,
  2007.

\end{thebibliography}

\end{document}